\documentclass[11pt]{article}
\usepackage{amsmath,amssymb}
\usepackage{arxiv_preprint}
\usepackage{booktabs,tabularx,graphicx}
\usepackage{flafter}
\usepackage{wrapfig}
\usepackage{caption}
\DeclareCaptionFont{compact}{\fontsize{8}{9.3}\selectfont}
\usepackage{needspace}
\usepackage{promptbox}
\usepackage{hyperref}
\usepackage{xurl}
\hypersetup{colorlinks=true, linkcolor=pplink, citecolor=pplink, urlcolor=pplink,
  pdftitle={Can LLMs Trust Appropriately? Evaluating Calibrated Reliance on Imperfect Partners},
  pdfauthor={Mengdi Chu; Jintao Huang}}
\renewcommand{\Needspace}[1]{}

\title{Can LLMs Trust Appropriately? \\ Evaluating Calibrated Reliance on Imperfect Partners}
\author{Mengdi Chu$^{1}$ \quad Jintao Huang$^{1}$}
\affiliations{$^{1}$The Ohio State University.}
\resources{\href{https://huggingface.co/datasets/90879c/llm-collaboration-trustbench/}{https://huggingface.co/datasets/90879c/llm-collaboration-trustbench}}

\begin{document}
\maketitle
\begin{abstract}
Large language models increasingly collaborate with humans and peer AI agents whose contributions may be incomplete, inconsistent, or incorrect. Effective collaboration requires calibrated trust, whereby decisions to rely on, verify, or assist a partner adapt rationally as empirical evidence accumulates. We investigate whether models translate partner evidence into appropriate reliance policies. We introduce a benchmark of 324 scenarios across 23 collaboration categories. Each scenario involves three sequential decisions with explicit transition dynamics, feedback emissions, and cost schedules, permitting exact analytical evaluation of choices against their expected consequences over the remaining interaction. Evaluating seven models alongside 36 controlled-history probes, we dissociate evidence sensitivity from normatively appropriate action. While the strongest frontier model approaches near-optimal decisions, other models exhibit recurring pathologies: procuring excessive safeguards, purchasing verification tests with zero expected value of information, prioritizing immediate savings over cooperative learning, and misallocating assistance to suboptimal channels. Supplying explicit posterior distributions resolves certain errors while introducing new failure modes. Calibrated trust in autonomous agents critically demands coupling empirical evidence with forward-looking decision utility.
\end{abstract}

\section{Introduction}
LLM agents increasingly collaborate with human users and peer agents by exchanging information, partitioning tasks, and reviewing mutual outputs \citep{wu2024autogen}. Effective cooperation requires strategic decisions regarding whose advice to follow, which tasks to delegate, and when to verify contributions. Accepting unreliable work risks propagating errors; conversely, exhaustive verification incurs prohibitive overhead and defeats delegation. Automation research has long formalized the dual hazards of excessive reliance and disuse \citep{parasuraman1997}. LLM collaborators face this foundational challenge of calibrating reliance on imperfect partners.

\begin{figure}[!t]
\centering
\includegraphics[width=\linewidth]{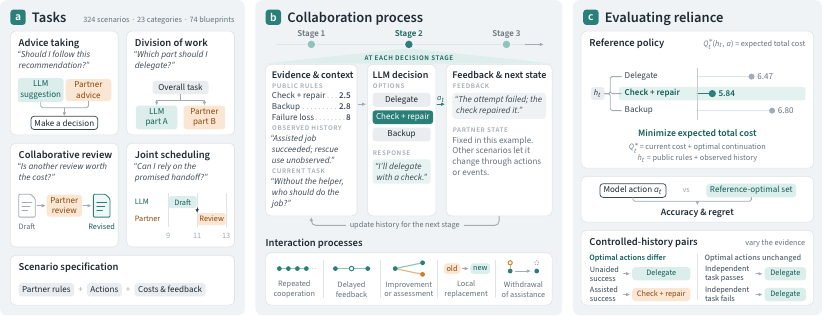}
\caption{Overview of our framework for evaluating how LLMs rely on imperfect partners. (a) The tasks cover 324 scenarios across 23 collaboration categories, with specified partner behavior, available actions, and costs. (b) Models make three sequential decisions. Their actions determine the feedback they receive and may affect future partner performance. (c) Decisions are compared with actions that minimize expected total cost given the available evidence. Controlled-history pairs examine whether models change or maintain their reliance appropriately as the evidence changes.}
\label{fig:benchmark_overview}
\end{figure}

Research on trust distinguishes perceived trustworthiness from the willingness to depend on a partner \citep{mayer1995}. Automation research connects trust to calibrated reliance within specific contexts \citep{lee2004}. In human--AI decision making, this entails accepting valid guidance while retaining independent judgment when advice is flawed \citep{schemmer2023}. We build on these foundations to investigate the behavioral expression of trust calibration in LLMs. Our central question is whether models translate partner evidence into normatively calibrated reliance, and under what conditions this translation fails.

Calibrated reliance depends on both accumulated evidence and the downstream task horizon. For example, success under guided assistance provides limited evidence regarding a partner's ability to operate autonomously. A reliable contribution may still warrant checking when error costs are severe, whereas inspecting an unreliable one yields negligible benefit if the result cannot alter the decision. Human reliance is sensitive to verification effort \citep{vasconcelos2023}. Sequential cooperation introduces dynamic dependencies: early actions shape both subsequent information and future partner performance. Delegation exposes partner strengths, takeover conceals them, and targeted assistance can actively enhance downstream performance. This interplay between reliance and adaptation is central to human--automation cooperation \citep{chiou2023}.

Recent studies show that LLM trust responds to partner characteristics and exhibits patterns reminiscent of human behavior \citep{xie2024,lerman2026}. \citet{chen2026} studies trust formation, breakdown, and recovery under costly verification protocols. Models can also account for speaker motives in controlled tasks, although this sensitivity is less consistent in naturalistic advertising contexts \citep{wu2025}. However, these findings leave open whether evidence-driven belief updates yield rational downstream decisions. An agent may react to partner failure by adopting heightened caution, yet select an unnecessarily costly safeguard, or purchase redundant tests while ignoring informative signals generated through routine collaboration. Evaluating these choices requires a normative reference policy that maps observed evidence to the costs and future consequences of candidate actions.

We introduce a benchmark of 324 scenarios across 23 collaboration categories (Figure~\ref{fig:benchmark_overview}). Each scenario formalizes partner generation processes, observational feedback, and cost structures over sequential stages. Models make three sequential decisions, enabling longitudinal evaluation of reliance dynamics as interactions unfold. Because transition rules and costs are fully specified, we can compute the Bayes-optimal reference policy and expected cost-to-go for every available action. An additional 36 pairs of controlled histories directly test whether models adapt reliance when warranted by evidence and maintain stability when the optimal action is invariant.

Evaluating seven models reveals near-optimal reliance in one frontier model (GPT-5.6 Sol) alongside systematic pathologies in others. Models frequently overreact to evidence by procuring excessive protection, purchase verification tests yielding zero expected value of information, or fail to exploit diagnostic feedback available through ordinary cooperation. Furthermore, assistance is frequently misdirected toward contributions whose improvement offers minimal downstream utility. Providing explicit posterior belief states rectifies certain errors while inducing novel failure modes. By coupling diverse cooperative workflows with analytical reference policies, our benchmark provides an objective testbed for evaluating whether empirical evidence translates into calibrated cooperative decisions.

\section{Background and Related Work}
Research in psychology and human--computer interaction (HCI) distinguishes perceived ability, benevolence, and integrity from the decision to depend on a partner \citep{mayer1995}, while automation research frames appropriate reliance in terms of the context and goals of use \citep{lee2004}. Studies in advice taking examine how people use others' judgments \citep{yaniv2007}, when they accept valid advice or reject errors \citep{schemmer2023}, and how verification effort modulates reliance choices \citep{vasconcelos2023}. Crucially, trust evolves dynamically with experience and circumstances \citep{hoff2015}. Trust repair depends on the type of violation and how the violating party responds \citep{kim2004}, while intrusive oversight can undermine collaborator effort \citep{falk2006}. These findings support a dynamic conception of trust grounded in adaptation and coordination \citep{chiou2023}, coupling reliance to accumulated evidence and the downstream effects of one's actions.

In LLM research, studies examine both model trustworthiness \citep{huang2024trustllm} and human reliance on AI \citep{zhou2025rel}. A complementary line investigates LLMs as trusting agents. \citet{xie2024} use trust games to compare LLM and human behavior, identifying partner-related biases. \citet{lerman2026} examine how competence, benevolence, integrity, and demographics shape LLM trust in humans. \citet{chen2026} operationalizes trust via costly verification against a memoryless baseline, analyzing formation, collapse, and repair.

Recent work explores how models exploit source metadata. \citet{wu2025} benchmark learning from motivated communication against rational models, finding sensitivity to motives improves when made salient. \citet{sun2026labels} show that both human and LLM judges assign higher trust to identical content labeled as human-authored than as AI-generated. \citet{cheng2026vigilance} examine how source credibility and framing affect vigilance against harmful beliefs, while \citet{pan2024credibility} integrate credibility into retrieval pipelines. We extend source evaluation to sequential cooperation, where relying, checking, and assisting incur tangible costs and shape downstream interactions.

\section{Benchmark Design}
\label{sec:benchmark}
Our benchmark evaluates whether LLMs translate empirical evidence about collaborators into normatively calibrated reliance. We operationalize calibrated reliance as choosing actions (e.g., delegation, verification, or assistance) that minimize cumulative expected cost under the scenario's fully specified rules. This decision-theoretic criterion connects an agent's evidentiary state with the normative validity of its actions.

\subsection{Interaction and reference decisions}
\label{sec:task_formulation}
Each scenario formalizes three sequential decision stages within a simulated environment. The initial prompt describes the setting, partner performance rules, candidate actions, and associated costs. At each stage, the model chooses an action (such as delegation, verification, assistance, or takeover) and receives the feedback that action reveals. Many environments involve two contribution channels, denoted P and Q, representing disparate partners, advisory sources, or modular sub-components with distinct performance rules. Collaborators are imperfect in that their contributions can fail, and reliability may vary across tasks or change over time. We formalize these properties via explicit transition dynamics.

Consider, for example, a sequential allocation task where an agent schedules jobs across domains $X$, $Y$, and $X$. Delegating execution reveals an observable outcome, whereas a paid backup guarantees completion without performance feedback. A software change after the first stage alters performance only in $X$, so the final choice hinges on the diagnostic persistence of the first outcome relative to the backup fee, while observations from $Y$ provide zero evidence about $X$ (Appendix~\ref{app:worked_example}).

We represent each scenario as a finite-horizon decision tree. Let $h_t$ contain public task specifications, past actions, and observed feedback, and let $A(h_t)$ be the admissible action set. Following the expected-cost formulation of partially observed decision problems \citep{kaelbling1998}, we compute:
\begin{equation}
\begin{aligned}
Q_t^*(h_t,a) &= \mathbb{E}[c_t\mid h_t,a]
 + \sum_o P(o\mid h_t,a)V_{t+1}^*(h_t,a,o),\\
V_t^*(h_t) &= \min_{a\in A(h_t)}Q_t^*(h_t,a),
\qquad V_4^*=0.
\end{aligned}
\label{eq:reference}
\end{equation}
Here, $c_t$ combines action fees and outcome costs, and $o$ is the next feedback observation. Rewards are represented as negative costs. Expectations integrate over latent partner states and future observation distributions given history $h_t$. The reference action set contains all minimizers of $Q_t^*(h_t,a)$, accounting for immediate costs and downstream information or state transitions under optimal continuation.

\subsection{Task construction and coverage}
\label{sec:construction}
The benchmark contains 324 scenarios spanning 23 categories of collaborative tasks across 74 blueprints. A category groups functionally related collaborative contexts; a blueprint defines the contribution mechanics and candidate action space. Each blueprint instantiates four or five scenarios by varying environmental parameters, cost schedules, and interaction processes (Appendix~\ref{app:inventory}).

We ground these coordination dilemmas in empirical literature on advice taking, trust repair, and behavioral responses to oversight \citep{yaniv2007,kim2004,falk2006}, as well as demonstration learning and collaborative grounding \citep{gweon2010,clark1986}. These empirical paradigms inform the operational challenges formalized in our benchmark: adjudicating conflicting advice, assessing uncertain evidence transfer across domains, diagnosing communication breakdowns, and tracking non-stationary partner capabilities. Each scenario specifies latent partner state distributions, observation emission functions, and action-cost schedules, yielding an analytically tractable reference policy (Appendix~\ref{app:design_sources}).

\Needspace{18\baselineskip}
\begin{wraptable}{r}{0.48\textwidth}
\vspace{-9pt}
\centering
\fontsize{8}{9.3}\selectfont
\captionsetup{font=compact,skip=5pt}
\caption{Interaction processes and scenario counts. Italics summarize the reliance decision. Appendix~\ref{app:inventory} lists the 23 collaboration categories.}
\label{tab:workflows}
\fontsize{8}{9.3}\selectfont
\setlength{\tabcolsep}{3pt}
\renewcommand{\arraystretch}{1.03}
\begin{tabularx}{\linewidth}{@{}Xr@{}}
\toprule
\textsf{Interaction process} & \textsf{$n$}\\
\midrule
\textsf{Repeated cooperation}\newline\textit{Learn from partner outcomes.} & 50\\[2pt]
\textsf{Delayed feedback}\newline\textit{Weigh earlier feedback against audit cost.} & 74\\[2pt]
\textsf{Improvement or assessment}\newline\textit{Choose between support and testing.} & 74\\[2pt]
\textsf{Local replacement}\newline\textit{Decide what to test before replacement.} & 74\\[2pt]
\textsf{Withdrawal of assistance}\newline\textit{Judge reliance after assistance ends.} & 28\\[2pt]
\midrule
\textsf{Specialized scenarios}\newline\textit{Contracts, disclosure, and other decisions.} & 24\\
\bottomrule
\end{tabularx}
\vspace{-5pt}
\end{wraptable}
Across 300 core scenarios, activity blueprints are crossed with five interaction processes governing feedback availability and partner state transitions (Table~\ref{tab:workflows}). These interaction topologies induce diverse strategic rationales for partner reliance. Certain tasks mandate conjunctive success across both channels, whereas others require only disjunctive sufficiency, where correlated failure modes constrain the risk reduction achieved through redundancy. Furthermore, actions differ in whether feedback isolates individual contributions or yields only an aggregated joint signal. Consequently, identical posterior reliability estimates can dictate divergent optimal actions across workflows. The remaining 24 specialized scenarios formalize coordination dilemmas involving contract design, asymmetric disclosure, strategic communication, and commitment.

All blueprints, task specifications, and environment rules were formulated and curated by the authors across structured domain rubrics. We exhaustively enumerate every decision tree to analytically verify optimal actions, transition probabilities, and continuation values along suboptimal branches (Appendix~\ref{app:inventory}).

In addition to the 324 primary scenarios, 36 pairs of controlled histories span all 23 categories. Each pair holds preceding action trajectories invariant while systematically varying the evidentiary history or the underlying data-generating process. Models execute a single decision per condition in isolated contexts. Optimal action sets are disjoint in 31 pairs and identical in five. These probes isolate evidence sensitivity from policy calibration, verifying whether models adapt reliance when warranted while maintaining stability when optimal policies remain invariant.

\section{Experimental Setup}
\label{sec:setup}
\subsection{Models and interaction protocol}
We evaluate seven models: GPT-6 Luna, GPT-5.6 Sol, Gemini 3.8 Flash, Claude Sonnet 5, Claude Haiku 4.5, DeepSeek V4.1 Flash, and Qwen3.8 27B. Prompts and the evaluation protocol are detailed in Appendix~\ref{app:protocol}.

All models receive identical task instructions and candidate actions, without external tools or auxiliary retrieval. Each scenario initializes in an isolated context window; subsequent turns append the cumulative history of public instructions, actions, and emitted feedback. Models return actions formatted as JSON. We evaluate one trajectory per primary scenario and one response per controlled-history input. Environmental feedback draws are pre-sampled and fixed at each node of the decision tree, ensuring strict comparability across models while allowing different policies to trace divergent paths.

On the 24 specialized scenarios, we also supply the exact Bayesian state distribution implied by the public rules and observed evidence. This condition isolates belief inference from policy selection, testing whether explicit reliability information aids models in selecting subsequent actions (Appendix~\ref{app:reliability_condition}).

\subsection{Evaluation measures}
\label{sec:metrics}
A decision is correct if it selects a reference-optimal action. We compute decision accuracy over reached stages within each blueprint and average these across all 74 blueprints. This blueprint-macro accuracy gives each blueprint equal weight. All-stage accuracy is the proportion of the 324 scenarios with correct actions across all three stages. Unparseable responses are scored as incorrect decisions. We also report overall decision accuracy, error counts, and unreached stages.

Decision regret is $g_t=Q_t^*(h_t,a_t)-V_t^*(h_t)$, the excess expected cost of an action under optimal continuation. To aggregate across heterogeneous cost scales, we normalize regret by the action-value spread at each node:
\begin{equation}
\widetilde g_t=\frac{g_t}{\max_{a\in A(h_t)}Q_t^*(h_t,a)-V_t^*(h_t)}.
\label{eq:normalized_regret}
\end{equation}
We average this quantity over reached decisions at nodes with a non-zero action-value spread.

For controlled histories, pair accuracy denotes the fraction of pairs where both decisions are optimal. We report agreement with the expected switch or stability relation as a secondary measure. We analyze performance across activity categories, interaction processes, and stages. Where actions follow a clear reliance ordering, we examine over-reliance and superfluous verification against reference actions and costs.

Targeted sub-analyses leverage task rules to assess whether test benefits exceed fees (positive EVOI), identify cases where greedy single-stage cost minimization conflicts with dynamic sequential optimality, and examine whether assistance targets the highest-value contribution. Because these subsets are defined a priori from reference value functions, they provide unconfounded evaluations of model behavior.

We report 95\% confidence intervals over task sampling, using bootstrap resampling of blueprints for aggregate metrics and Wilson score intervals for diagnostic proportions (Appendix~\ref{app:uncertainty}).

\Needspace{9\baselineskip}
\section{Results and Analysis}
\label{sec:results}
Blueprint-macro decision accuracy ranges from 80.9\% to 99.0\%, and all-stage accuracy from 56.5\% to 96.9\% (Table~\ref{tab:main_results_v5}). Sol selects reference-optimal actions across all three stages in 314 of 324 scenarios (96.9\%), demonstrating that sustained, calibrated reliance is achievable within these environments. Below, we examine how models interpret partner evidence, arbitrate verification decisions, learn through cooperation, allocate assistance, and respond to explicit reliability distributions.

\begin{table}[htbp]
\centering\small
\caption{Overall performance on the trust calibration benchmark (\%). Decision accuracy averages over 74 blueprints. All-stage accuracy requires optimal choices at all three stages (324 scenarios); paired accuracy requires optimal choices in both histories (36 pairs). Normalized regret measures excess expected cost over reached decisions, scaled by the range of action values. Brackets show 95\% confidence intervals, using blueprint bootstrap for main-task metrics and Wilson intervals for paired accuracy. Decision counts are reported in Appendix~\ref{app:outcomes}.}
\label{tab:main_results_v5}
\begingroup
\fontsize{9}{10.5}\selectfont
\setlength{\tabcolsep}{3pt}
\renewcommand{\arraystretch}{1.15}
\newcommand{\resultcell}[3]{\makebox[2.6em][r]{#1}\hspace{3pt}{\fontsize{7.5}{8.5}\selectfont\color[gray]{0.38}\makebox[5.9em][l]{[#2, #3]}}}
\begin{tabular*}{\linewidth}{@{}l@{\extracolsep{\fill}}rrrr@{}}
\toprule
Model & \multicolumn{1}{c}{Decision acc. $\uparrow$} & \multicolumn{1}{c}{All-stage acc. $\uparrow$} & \multicolumn{1}{c}{Norm. regret $\downarrow$} & \multicolumn{1}{c@{}}{Paired acc. $\uparrow$}\\[1pt]
\midrule
GPT-6 Luna & \resultcell{91.0}{89.3}{92.7} & \resultcell{74.7}{70.0}{79.3} & \resultcell{2.96}{2.24}{3.72} & \resultcell{80.6}{65.0}{90.2}\\[1pt]
GPT-5.6 Sol & \resultcell{99.0}{98.4}{99.6} & \resultcell{96.9}{94.8}{98.8} & \resultcell{0.22}{0.07}{0.40} & \resultcell{100.0}{90.4}{100.0}\\[1pt]
Gemini 3.8 Flash & \resultcell{93.2}{91.6}{94.6} & \resultcell{81.5}{77.7}{85.2} & \resultcell{3.72}{2.79}{4.69} & \resultcell{97.2}{85.8}{99.5}\\[1pt]
Claude Sonnet 5 & \resultcell{85.0}{82.9}{87.2} & \resultcell{62.3}{57.2}{67.4} & \resultcell{6.68}{5.48}{7.95} & \resultcell{86.1}{71.3}{93.9}\\[1pt]
Claude Haiku 4.5 & \resultcell{80.9}{77.6}{84.3} & \resultcell{58.3}{51.9}{64.8} & \resultcell{7.99}{6.51}{9.50} & \resultcell{83.3}{68.1}{92.1}\\[1pt]
DeepSeek V4.1 Flash & \resultcell{90.1}{86.8}{92.9} & \resultcell{77.8}{72.4}{82.9} & \resultcell{1.19}{0.67}{1.79} & \resultcell{94.4}{81.9}{98.5}\\[1pt]
Qwen3.8 27B & \resultcell{81.0}{78.4}{83.7} & \resultcell{56.5}{51.1}{62.0} & \resultcell{5.19}{3.90}{6.53} & \resultcell{80.6}{65.0}{90.2}\\[1pt]
\bottomrule
\end{tabular*}
\endgroup
\end{table}

\subsection{Responding to partner evidence}
\label{sec:evidence_results}
\begin{figure}[htbp]
\centering
\includegraphics[width=\linewidth]{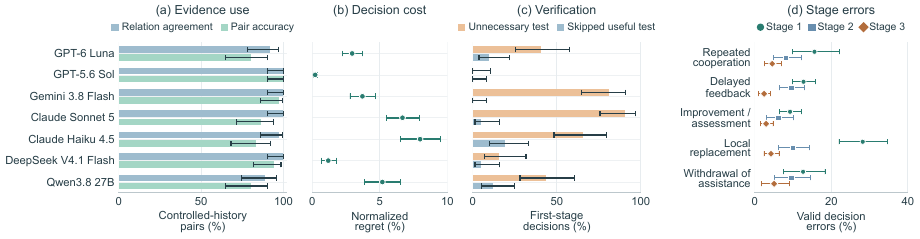}
\caption{Trust calibration in collaborative decisions. (a) Relation agreement and paired accuracy under controlled partner histories (36 pairs). (b) Mean normalized regret over reached decisions. (c) Unnecessary tests ($n=32$) and skipped useful tests ($n=42$). (d) Error rates among reached decisions, pooled across models by interaction process and stage. Panels (a--c) share model rows. Error bars show 95\% Wilson intervals in (a, c) and blueprint-bootstrap intervals in (b, d).}
\label{fig:reliance_results}
\end{figure}

Gemini, Sonnet, and DeepSeek match the expected switch or stability relation in all 36 controlled-history pairs (100\%), yet their paired accuracies drop to 97.2\%, 86.1\%, and 94.4\%, respectively (Figure~\ref{fig:reliance_results}a). Relation agreement verifies whether choices switch or remain stable; paired accuracy requires an optimal action across both conditions. Across models, agreement without paired correctness manifests in nine scenarios spanning seven categories, including verification, knowledge allocation, and information sharing (Appendix~\ref{app:controlled_results}).

Contract selection illustrates the behavioral hazard of this gap. Following partner success, retaining the baseline contract minimizes expected cost; after failure, transitioning to an incentive-aligned performance contract is optimal. Gemini and Sonnet instead deploy the performance contract after success and escalate to a guaranteed contract after failure. While their choices adapt to evidence, both models procure excessive protection relative to normative requirements. Assessing trust calibration therefore requires evaluating not merely whether models respond to evidence, but whether the economic cost and form of the resulting reliance are calibrated.

\subsection{Deciding when to verify}
\label{sec:verification_results}
In local replacement, models can audit either of two collaborator channels before execution; one channel is subsequently subject to potential replacement. For all 74 scenarios, models receive identical initial inputs. Foregoing verification is optimal in 32 scenarios spanning 19 activity categories, whereas testing is optimal in the remaining 42.

\begin{figure}[!t]
\centering
\includegraphics[width=\linewidth]{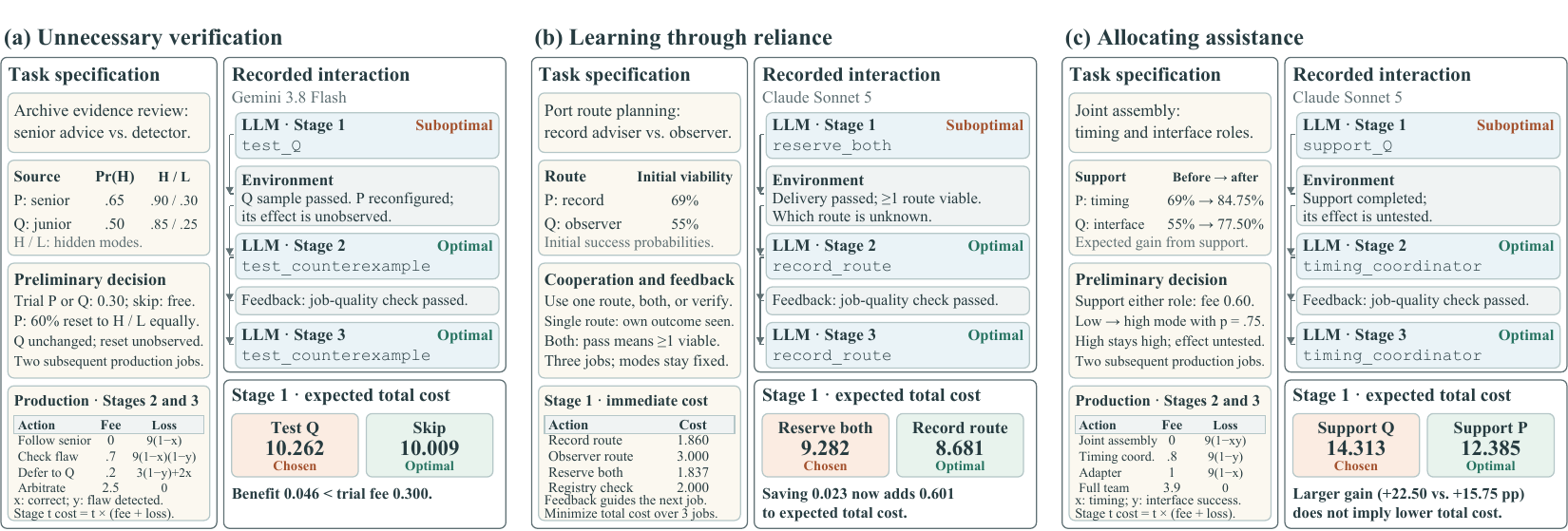}
\caption{Examples of reliance decisions across three collaboration scenarios. (a) A trial provides useful information, but its expected benefit falls below its cost. (b) Reserving both routes saves immediate cost but limits learning that would improve later decisions. (c) Supporting the channel with the larger reliability gain leads to higher expected cooperation cost. Costs compare initial actions under optimal continuation.}
\label{fig:reliance_cases}
\end{figure}

Six models persistently acquire superfluous tests (Figure~\ref{fig:reliance_results}c), with Gemini doing so in 26/32 cases (81.3\%) and Sonnet in 29/32 (90.6\%). Strikingly, of the 108 unnecessary tests across models, 107 target the unchanged channel. While testing an invariant collaborator ensures evidentiary persistence, its cost strictly exceeds its downstream informational value.

A substantial portion of tests provides zero expected value of information (EVOI). In 17 scenarios, neither candidate test reduces downstream costs even before deducting fees, assuming optimal continuation. Gemini nevertheless tests in 12 of these scenarios and Sonnet in 15. Across all 108 unneeded tests, 61 offer zero gross benefit prior to fees (Appendix~\ref{app:acquisition}). In these settings, feedback merely refines posterior beliefs without any capacity to alter the subsequent decision.

Figure~\ref{fig:reliance_cases}a highlights an instance where information value is positive yet economically insufficient. An archive team tests a junior counterexample source for $0.30$, reducing downstream expected loss by merely $0.046$. While Sol skips verification, the remaining six models test, inflating expected total cost from $10.009$ to $10.262$ under optimal continuation. Gemini subsequently observes a successful trial and executes optimal downstream choices, illustrating that downstream competence can easily mask an economically irrational upstream verification choice.

Models also miss valuable evidence. Haiku skips a worthwhile test in 8/42 cases (19.0\%), and Qwen in 5/42 (11.9\%). Both failure modes, purchasing sub-economic evidence and neglecting profitable audits, substantially degrade collaborative efficiency.

These initial choices account for 142 of the decision errors in local replacement. The pooled first-stage error rate is 28.2\% (95\% CI 22.0--34.6), markedly higher than the 10.1\% and 4.2\% observed at subsequent stages along reached trajectories (Figure~\ref{fig:reliance_results}d). The primary vulnerability in this interaction topology lies in deciding whether to acquire diagnostic information before collaboration begins.

\Needspace{18\baselineskip}
\subsection{Learning about partners through cooperation}
\label{sec:learning_results}
\begin{wraptable}{r}{0.49\textwidth}
\vspace{-9pt}
\centering
\fontsize{8}{9.3}\selectfont
\captionsetup{font=compact,skip=5pt}
\caption{First choices in 15 scenarios where learning outweighs immediate savings. Long-term choices minimize three-stage expected cost; immediate choices minimize current expected cost (Appendix~\ref{app:information_planning}).}
\label{tab:information_planning}
\fontsize{8}{9.3}\selectfont
\setlength{\tabcolsep}{3pt}
\renewcommand{\arraystretch}{1.08}
\begin{tabular*}{\linewidth}{@{}l@{\extracolsep{\fill}}rr@{}}
\toprule
Model & Long-term $\uparrow$ & Immediate $\downarrow$\\
\midrule
GPT-6 Luna & 12/15 & 3/15\\[1pt]
GPT-5.6 Sol & 15/15 & 0/15\\[1pt]
Gemini 3.8 Flash & 12/15 & 3/15\\[1pt]
Claude Sonnet 5 & 1/15 & 14/15\\[1pt]
Claude Haiku 4.5 & 0/15 & 15/15\\[1pt]
DeepSeek V4.1 Flash & 9/15 & 3/15\\[1pt]
Qwen3.8 27B & 3/15 & 8/15\\[1pt]
\bottomrule
\end{tabular*}
\vspace{-5pt}
\end{wraptable}

Cooperation itself serves an exploratory function by generating diagnostic feedback about partner reliability. Among the 50 repeated-cooperation scenarios, 15 (spanning 13 categories) embody an active information-planning trade-off: the optimal first action incurs a higher immediate expected cost to obtain feedback that improves downstream decisions. These cases are identified analytically from reference values, selecting scenarios where no greedy single-stage cost minimizer is optimal over the remaining rounds. Haiku chooses the myopic cost minimizer in all 15 cases, and Sonnet does so in 14 (Table~\ref{tab:information_planning}). In contrast, Sol selects the optimal exploratory action in all 15; Luna and Gemini do so in 12 each.

A route-planning scenario illustrates this dynamic exploration--exploitation trade-off (Figure~\ref{fig:reliance_cases}b). Concurrently reserving both advisers' routes incurs an immediate expected cost of $1.837$, marginally lower than relying on the record adviser alone ($1.860$). However, single-channel delegation isolates whether that adviser was viable, whereas joint reservation reveals only whether at least one route succeeded. Under optimal continuation, absorbing the minor initial cost penalty to isolate feedback yields an expected cumulative cost of $8.681$, compared with $9.282$ under immediate hedging. Sonnet and Haiku greedily choose the latter, securing marginal immediate savings while forfeiting critical diagnostic feedback.

Model behavior diverges sharply when identical diagnostic information is packaged as an explicit audit rather than an emergent byproduct of work allocation. Seven of the 15 blueprints admit matched delayed-feedback counterparts where purchasing an audit is optimal. These workflows share production payoffs and prior state probabilities, differing only in feedback timing and available actions. Luna makes optimal first choices in 4/7 ordinary-cooperation cases and 5/7 audit cases; Gemini does so in 5/7 and 7/7, respectively (Figure~\ref{fig:partner_information}a). Sonnet displays the starkest contrast, choosing the myopic cost minimizer in all seven ordinary-cooperation cases while buying the optimal audit in all seven audit cases (across all seven pairs; Appendix~\ref{app:information_planning}).

This contrast separates two modalities of partner learning: executing dedicated verification actions versus structuring routine task division to harvest diagnostic signals. Sonnet readily pays for explicit audits yet consistently overlooks the informational dividends of cooperative task allocation. Coupled with its unneeded tests in Section~\ref{sec:verification_results}, this dissociation demonstrates why verification frequency alone is an inadequate proxy for trust calibration: true calibration requires aligning information acquisition with cumulative decision utility.

\subsection{Allocating assistance for future cooperation}
\label{sec:dynamic_results}
In improvement-or-assessment scenarios, models can support or test either channel, or do neither. Providing assistance enhances future capability, but the optimal target depends on how channels are leveraged in downstream work.

Several models consistently misallocate support interventions toward suboptimal channels (Figure~\ref{fig:partner_information}b). Of Sonnet's nine first-stage errors, eight target channel Q when assisting P is optimal; Gemini makes this error in five of its seven first-stage errors. In an assembly task where channel Q begins with lower baseline reliability and offers a larger nominal gain from assistance (Figure~\ref{fig:reliance_cases}c), equal intervention fees nevertheless favor improving P due to its downstream structural importance ($12.385$ vs. $14.313$ for Q). Sonnet erroneously assists Q, focusing on local reliability improvements while neglecting downstream utility. Appendix~\ref{app:assistance} details the full error breakdown and analytical derivations.

\subsection{Using explicit reliability information}
\label{sec:reliability_information}
\begin{figure}[htbp]
\centering
\includegraphics[width=\linewidth]{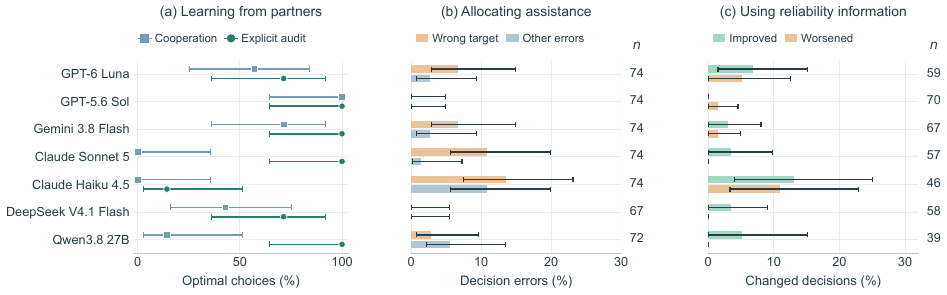}
\caption{Learning about partners and supporting cooperation. (a) Optimal first choices in ordinary-cooperation and explicit-audit workflows on seven shared blueprints. (b) Error rates among initial improvement-or-assessment decisions. Wrong target means supporting one channel when supporting the other is optimal. (c) Changes after supplying partner-state probabilities, comparing decisions with identical preceding actions and feedback. Improved denotes incorrect-to-correct changes, and worsened the reverse. Model rows are shared; $n$ gives each panel's denominator. Error bars show 95\% Wilson intervals in (a, b) and blueprint-bootstrap intervals in (c).}
\label{fig:partner_information}
\end{figure}

Does isolating belief inference from decision-making improve cooperation? On 24 specialized scenarios, we explicitly supply the normative joint distribution over partner states, derived from public task rules and observed history. This provides exact Bayesian belief updates while leaving action selection to the model. Conditioning on identical historical trajectories yields 39--70 paired decisions per model (Figure~\ref{fig:partner_information}c).

Most evaluated decisions are already optimal in both conditions. Performance gains on previously failed instances are largely offset by novel mistakes. Luna rectifies four of its six baseline errors but changes three previously optimal decisions to incorrect choices. Similarly, Haiku repairs six of eight errors while introducing five new mistakes, netting an overall gain of only one additional optimal decision across 59 and 46 paired decisions, respectively (Figure~\ref{fig:partner_information}c, Appendix~\ref{app:reliability_support}).

Providing explicit posterior belief states remediates isolated misjudgments, yet its net workflow utility depends equally on preserving baseline competence. Evaluating decision support must track both error rectification and induced degradation, as counting repairs in isolation overstates cooperative gains.

\section{Discussion}
\label{sec:discussion}
An agent adopting heightened caution after partner failure may appear to calibrate trust, yet choose safeguards whose cost exceeds their benefit. Decoupling responsiveness to evidence from calibrated reliance is therefore essential. The same principle governs assistance: improving partner capability is justified only when it yields downstream collaborative payoff.

Cooperation also governs what an agent can infer. When an action isolates individual contributions, the resulting evidence sharpens future delegation; when joint outcomes aggregate contributions, credit assignment becomes harder. Work allocation thus fulfills an informational role alongside immediate production. The contrast between ordinary cooperation and explicit audits demonstrates that learning through work represents a distinct cognitive demand from willingness to verify: structuring informative workflows is an integral facet of trust calibration.

Decision support must be evaluated by both errors corrected and competencies preserved. Supplied distributions resolve isolated errors, but repairs and new mistakes occur together. A valuable next step is investigating scaffolding that directly connects candidate actions to expected costs, feedback distributions, and downstream consequences, testing which interventions improve cooperation while preserving existing strengths.

\section{Conclusion}
\label{sec:conclusion}
We introduced a benchmark of 324 scenarios across 23 collaboration categories to investigate whether LLMs translate partner evidence into appropriate reliance. Our seven-model evaluation reveals that responsiveness to evidence frequently coexists with excessive protection, superfluous verification with zero expected value of information, and missed opportunities to learn through cooperation. Assistance is often misdirected toward suboptimal targets, while explicit reliability distributions induce new failure modes alongside targeted repairs. Calibrated trust cannot be reduced to reactive caution or inspection frequency; it requires forward-looking coordination between empirical evidence, operational costs, and downstream task horizons.

\subsection*{AI use statement}
Generative AI tools assisted with literature discovery, code development, data analysis, and manuscript preparation. The authors are responsible for the study design, analysis, interpretation of results, and final manuscript.

\subsection*{Reproducibility statement}
The benchmark and code are available at \url{https://huggingface.co/datasets/90879c/llm-collaboration-trustbench/}. Appendices~\ref{app:inventory} and~\ref{app:methods} describe task construction, reference calculations, model configurations, and scoring. The release includes the evaluation environments, prompts, feedback draws, and the code for running and scoring models on the benchmark.

\bibliographystyle{plainnat}
\bibliography{references}

\appendix
\let\ppclearpage\clearpage
\renewcommand{\clearpage}{\par}
\section{Task Inventory and Construction Details}
\label{app:inventory}
Table~\ref{tab:inventory} lists the 23 collaboration categories. A category groups related activities; a blueprint specifies how contributions combine and which actions are available; a scenario instantiates a blueprint with a setting, costs, and an interaction process. The 74 blueprints contribute four or five scenarios each, giving 324 scenarios in total. Of these, 300 use the five interaction processes in Table~\ref{tab:workflows}, and 24 use specialized environments for contracts, disclosure, communication, and other forms of cooperation.

\begin{table}[ht]
\centering\small
\caption{Activity coverage of the benchmark.}
\label{tab:inventory}
\begin{tabularx}{\linewidth}{@{}lXrr@{}}
\toprule
ID & Activity category & Blueprints & Scenarios\\
\midrule
F01 & Advice conflict resolution & 4 & 17\\
F02 & Numerical advice integration & 2 & 9\\
F03 & Collaborative plan review and repair & 6 & 26\\
F04 & Exploration after demonstration & 3 & 13\\
F05 & Verification and help seeking & 2 & 9\\
F06 & Task allocation and delegation & 4 & 18\\
F07 & Allocation of review resources & 5 & 22\\
F08 & Monitoring and takeover & 3 & 13\\
F09 & Alarm response & 2 & 9\\
F10 & Integration of multiple evidence sources & 4 & 18\\
F11 & Team knowledge retrieval and referral & 3 & 13\\
F12 & Entrustment and fulfillment & 3 & 13\\
F13 & Coordinated investment and commitments & 5 & 22\\
F14 & Authorization and responsibility boundaries & 4 & 17\\
F15 & Entrusted information sharing & 2 & 9\\
F16 & Trials and restoration of reliance & 3 & 13\\
F17 & Shared understanding and communication repair & 3 & 13\\
F18 & Contracts and behavioral responses & 2 & 9\\
F19 & Partner selection and replacement & 3 & 13\\
F20 & Voicing concerns and communicating errors & 2 & 9\\
F21 & Joint scheduling and shared resources & 4 & 17\\
F22 & Teaching and withdrawal of support & 2 & 9\\
F23 & Distributed search and coverage handoffs & 3 & 13\\
\midrule
 & Total & 74 & 324\\
\bottomrule
\end{tabularx}
\end{table}

Task diversity encompasses both the structural topology of collaborative contributions and the dynamic interaction processes governing execution. For example, parallel routing provides modular redundancy, whereas sequential assembly requires conjunctive success across all participating channels. Feedback mechanisms vary between unconfounded channel-specific observations and aggregated joint outcomes, while downstream stages may introduce partner transitions or assistance withdrawal. These structural variations critically modulate what an agent can infer and dictate optimal work allocation.

To describe structural overlap, we also group blueprints by the ordering of outcome costs within each action across possible contribution outcomes. This comparison ignores action labels, cost magnitudes, action fees, and temporal processes, yielding 31 groups from the 74 blueprints. The 31 groups summarize static outcome relations; the complete scenarios additionally specify feedback, state changes, and sequential choices. We aggregate performance by blueprint and report separate breakdowns by collaboration category and interaction process.

Reference calculations enumerate all legal actions and feedback branches. Offline checks verify probability normalization, expected costs, optimal-action ties, and continuation after legal suboptimal choices. The environments contain 10,652 decision nodes, 43,086 node-action options, and 75,772 feedback branches. Comparing an optimal feedback-dependent policy with the best fixed sequence identifies 277 scenarios with positive adaptation value and 47 with zero adaptation value. Both groups were determined from the reference policies before model evaluation.

\Needspace{9\baselineskip}
\subsection{From cooperation research to benchmark tasks}
\label{app:design_sources}
Foundational literature in organizational psychology, behavioral economics, and cognitive science informs the core cooperative dilemmas formalized in our benchmark. Each theoretical paradigm is translated into an executable decision environment by operationalizing contribution channels, latent partner states, observational feedback functions, and action-consequence topologies. Table~\ref{tab:design_sources} connects four foundational research paradigms to the resulting tasks and trust calibration questions. The numerical parameters and reference policies belong to the constructed environments.

\begin{table}[htbp]
\centering\small
\caption{Connections between cooperation research, task blueprints, and the trust-related decisions they examine. Each blueprint is instantiated within the interaction processes described in the main text.}
\label{tab:design_sources}
\begin{tabularx}{\linewidth}{@{}>{\raggedright\arraybackslash}p{0.23\linewidth}>{\raggedright\arraybackslash}X>{\raggedright\arraybackslash}p{0.26\linewidth}@{}}
\toprule
Research basis & Benchmark decision & Trust-related question\\
\midrule
Advice taking \citep{yaniv2007} & Choose one adviser's shipping route, reserve both routes, or buy a definitive check (F01.T1). & Which adviser should guide the choice, and what can that choice reveal?\\[5pt]
Sampling and generalization \citep{gweon2010} & Use a selected demonstration or a randomly sampled practice example for a fabrication task (F04.T3). & When does demonstrated performance support reliance on a new task?\\[5pt]
Collaborative reference \citep{clark1986} & Reuse local terminology, clarify it with one recipient, or use explicit references with both (F17.T1). & Which partner shares the understanding needed for successful work?\\[5pt]
Behavioral responses to control \citep{falk2006} & Choose discretion, detailed control, mixed roles, or external execution (F18.T2). & How should reliance account for performance under different forms of control?\\
\bottomrule
\end{tabularx}
\end{table}

\section{Task and Evaluation Details}
\label{app:methods}
\subsection{A worked example}
\label{app:worked_example}
Below is the full first-turn user input for scenario \texttt{U5.F01.T2.A}. The system prompt is given in Appendix~\ref{app:protocol}.

\begin{promptbox}[breakable]{User prompt: initial task input (\texttt{U5.F01.T2.A})}
This is a simulated collaboration task. Use only the stated rules and actual feedback. Do not use tools or invent observations. No running score is shown; unreported outcomes remain unknown. Return one JSON object for the current decision, without future-stage answers.\par\smallskip
Complete three jobs in order: a small X job, a Y job, then a final X job.\par\smallskip
The partner has a separate hidden G/W state for each domain. Initially the states are independent, with P(G)=0.70 in each. Within a domain, success is 0.90 in G and 0.30 in W. Given current states, fresh performance draws are independent.\par\smallskip
For each job choose delegate, which is free and reports actual success/failure after work, or backup, which guarantees completion but gives no partner-performance observation. Failure losses for the three jobs are 4, 8, 12; backup fees are 1.8, 2.6, 3.5 respectively.\par\smallskip
After round 1, a software change affects X only: with probability 0.40, its X state is freshly redrawn as G/W with equal probabilities; otherwise that state stays unchanged. This happens whether you delegated or used backup, is independent of earlier outcomes, and is not directly observed. Y does not change. There are no later state changes.\par\smallskip
Minimize total expected fees and failure losses across all three jobs.\par\smallskip
Round 1: Complete the small X job.\par\smallskip
Return \texttt{\{"action":"delegate/backup"\}} for this decision only.
\end{promptbox}

If the model delegates and the partner succeeds, the next message is:

\begin{promptbox}{Environment feedback after a successful first job}
Partner job succeeded. The scheduled X software change has now occurred; its effect is unobserved.\par\smallskip
Round 2: Complete the Y job.\par\smallskip
Return \texttt{\{"action":"delegate/backup"\}} for this decision only.
\end{promptbox}

The reference solution is not shown to the model. A success raises $P(G_X)$ from $0.70$ to $0.875$. After the software change it becomes $0.6(0.875)+0.4(0.5)=0.725$, so the final $X$ job succeeds with probability $0.9(0.725)+0.3(0.275)=0.735$. Delegating that job then has expected loss $12(1-0.735)=3.18$, less than the backup fee of $3.5$. After a failure, $P(G_X)$ is $0.25$, then $0.35$ after the change; the success probability drops to $0.51$ and delegation costs $5.88$, so backup is better. The $Y$ outcome changes neither case. At the first decision, delegating has expected total cost $6.63$ and backup $7.54$, so delegation is optimal.

\subsection{Prompts and evaluation protocol}
\label{app:protocol}
All conditions use the same system prompt:
\begin{promptbox}{System prompt}
You are a language model participating in a controlled collaboration decision experiment. Answer using the current task and actual feedback, following the requested output format.

Answer using only the experimental materials supplied in each input. Do not use tools or access files or the network. Follow the requested output format.
\end{promptbox}
Each scenario runs in a fresh context, and later turns replay the visible conversation. The prompt asks for JSON, and we parse the action from the response, including JSON inside prose or a code block. Any legal action, optimal or not, moves the environment down the corresponding branch. An empty, ambiguous, or unparseable response ends the trajectory and counts as an error.

\subsection{Scoring and uncertainty}
\label{app:scoring}
\label{app:uncertainty}
An action is correct if its expected cost matches the optimum; when several actions tie, each receives credit. Normalized regret skips nodes where all actions have the same value. Blueprint-macro accuracy is the mean of the 74 per-blueprint accuracies. For a controlled-history pair, paired accuracy requires both actions to be optimal, while relation agreement only checks whether the action changes or stays the same as it should.

Confidence intervals for the main metrics come from a bootstrap over blueprints (10,000 resamples), keeping all scenarios of a blueprint together and using the same resamples for every model. For proportions in the diagnostic analyses we use 95\% Wilson intervals.

\subsection{Reliability-support condition}
\label{app:reliability_condition}
In this condition, every decision request also includes the exact posterior over the partner's hidden states, computed from the task rules and the observations so far. In the example of Appendix~\ref{app:worked_example}, after a successful first job the message also contains:

\begin{promptbox}{Reliability support appended to the prompt}
Belief support, calculated only from the public task rules and the observations so far (probabilities rounded to 10 decimal places):\par\smallskip
X=G, Y=G: 0.5075000000\\
X=G, Y=W: 0.2175000000\\
X=W, Y=G: 0.1925000000\\
X=W, Y=W: 0.0825000000\\
This is a distribution over possible current states, not a revelation of the actual state.
\end{promptbox}

This corresponds to $P(G_X)=0.725$ and $P(G_Y)=0.7$. It gives no action values or recommendation. We run the 24 specialized scenarios this way with the same feedback draws as the main runs, and compare decisions only at stages where both runs share the same history and both responses are valid.

\subsection{Controlled histories requiring a changed or stable choice}
\label{app:history_examples}
Each controlled-history input gives the task rules and a fixed history, then asks for one decision. The two histories in a pair run in separate contexts. We show two pairs below; the optimal actions are noted for the reader and are not shown to the model.

\textit{A changed choice.}
In \texttt{U5.F18.T1.A}, the partner works under either a default contract or a performance contract, which costs more but can elicit extra effort. Both histories start with the default contract, and the prompt includes:
\begin{promptbox}{Diagnostic instructions}
This is a controlled-history diagnostic. The round-1 action was assigned for this case; do not judge or redo it. Treat its observation as given. Earlier costs are sunk. Optimize the remaining objective. Supply only the next decision.
\end{promptbox}
\noindent\begin{minipage}[t]{0.485\linewidth}
\begin{promptbox}{History A}
Round 1: action=default\_contract. Observation: Job-quality check failed.
\end{promptbox}
\end{minipage}\hfill
\begin{minipage}[t]{0.485\linewidth}
\begin{promptbox}{History B}
Round 1: action=default\_contract. Observation: Job-quality check passed.
\end{promptbox}
\end{minipage}
\par\medskip
The model then chooses among the default contract, the performance contract, an audit alone, or a guaranteed contract for the two remaining jobs. After a failed check, \texttt{performance\_contract} is optimal; after a passed check, \texttt{default\_contract} is.

\Needspace{15\baselineskip}
\textit{A stable choice.}
This pair uses the task from Appendix~\ref{app:worked_example}. Both histories delegate the first two jobs and share the first observation:
\begin{promptbox}{Shared history: round 1}
Round 1: action=delegate. Observation: Partner job succeeded. The scheduled X software change has now occurred; its effect is unobserved.
\end{promptbox}
\noindent\begin{minipage}[t]{0.485\linewidth}
\begin{promptbox}{History A}
Round 2: action=delegate. Observation: Partner job succeeded.
\end{promptbox}
\end{minipage}\hfill
\begin{minipage}[t]{0.485\linewidth}
\begin{promptbox}{History B}
Round 2: action=delegate. Observation: Partner job failed.
\end{promptbox}
\end{minipage}
\par\medskip
The model must decide whether to delegate the final $X$ job. The second observation concerns $Y$, which is independent of $X$, so the final $X$ success probability stays at $0.735$ in both histories. Delegation (expected cost $3.18$) beats backup ($3.5$) in both, so \texttt{delegate} is optimal regardless of the $Y$ outcome.

\clearpage
\section{Additional Results}
\label{app:results_detail}
\subsection{Decision outcomes and activity categories}
\label{app:outcomes}
\begin{figure}[p]
\centering
\includegraphics[width=\linewidth]{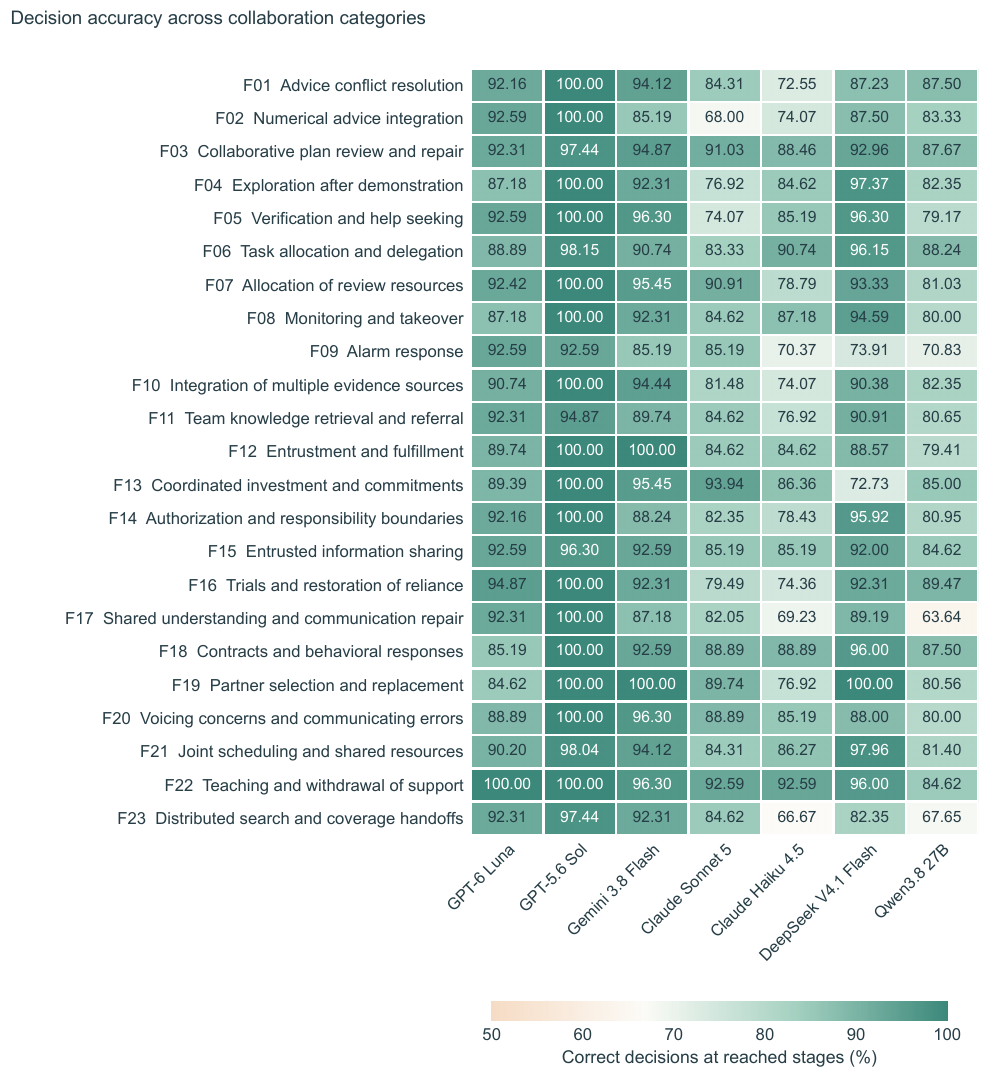}
\caption{Decision accuracy across the 23 collaboration categories. Cells show the percentage of optimal actions among reached stages. Model names and order match Table~\ref{tab:main_results_v5}.}
\label{fig:category_results}
\end{figure}

Table~\ref{tab:output_validity} reports reached stages, correct actions, decision errors, and stages that are not reached. Normalized regret in Table~\ref{tab:main_results_v5} is evaluated over reached decisions, whereas decision accuracy gives each blueprint equal weight. Figure~\ref{fig:category_results} reports accuracy within each of the 23 collaboration categories. These category scores pool reached stages, whereas the main table averages accuracy across blueprints.

\begin{table}[htbp]
\centering\small
\caption{Main-scenario decision counts. Trajectories terminate upon an unparseable response, and subsequent stages are recorded as unreached.}
\label{tab:output_validity}
\setlength{\tabcolsep}{5pt}
\renewcommand{\arraystretch}{1.1}
\begin{tabular}{@{}lrrrr@{}}
\toprule
Model & Reached stages & Correct & Decision errors & Unreached\\
\midrule
GPT-6 Luna & 972 & 884 & 88 & 0\\
GPT-5.6 Sol & 972 & 962 & 10 & 0\\
Gemini 3.8 Flash & 972 & 906 & 66 & 0\\
Claude Sonnet 5 & 970 & 826 & 144 & 2\\
Claude Haiku 4.5 & 972 & 787 & 185 & 0\\
DeepSeek V4.1 Flash & 890 & 813 & 77 & 82\\
Qwen3.8 27B & 874 & 716 & 158 & 98\\
\bottomrule
\end{tabular}
\end{table}

\subsection{Controlled-history decisions}
\label{app:controlled_results}
Table~\ref{tab:evidence_pairs} separates the 31 pairs requiring a switch from the five requiring stability. Relation agreement requires both decisions to exhibit the expected switch or stability relation; pair accuracy additionally requires both actions to be optimal. Appendix~\ref{app:history_examples} shows one input pair of each kind.

The contract example in Section~\ref{sec:evidence_results} is \texttt{U5.F18.T1.A}. Following the assigned default contract, the optimal next choice is the performance contract after failure and the default contract after success. Both Gemini and Sonnet instead choose the guaranteed contract after failure and the performance contract after success.

Relation agreement without paired correctness occurs in 20 model--pair observations involving nine scenarios across seven categories. All six models other than Sol show this mismatch in the contract pair. Other cases concern checking, workload, demonstration transfer, knowledge allocation, disclosure, and search.

\begin{table}[htbp]
\centering\small
\caption{Controlled-history results. Switch and stable columns count pairs with both actions correct; relation correctness evaluates only whether actions change or stay the same.}
\label{tab:evidence_pairs}
\setlength{\tabcolsep}{3.5pt}
\renewcommand{\arraystretch}{1.1}
\begin{tabular}{@{}lrrrr@{}}
\toprule
Model & \shortstack{Both correct\\(/36)} & \shortstack{Switch pairs\\(/31)} & \shortstack{Stable pairs\\(/5)} & \shortstack{Relation correct\\(/36)}\\
\midrule
GPT-6 Luna & 29 & 24 & 5 & 33\\
GPT-5.6 Sol & 36 & 31 & 5 & 36\\
Gemini 3.8 Flash & 35 & 30 & 5 & 36\\
Claude Sonnet 5 & 31 & 26 & 5 & 36\\
Claude Haiku 4.5 & 30 & 26 & 4 & 35\\
DeepSeek V4.1 Flash & 34 & 29 & 5 & 36\\
Qwen3.8 27B & 29 & 25 & 4 & 32\\
\bottomrule
\end{tabular}
\end{table}

\clearpage
\clearpage

\subsection{Information acquisition before local replacement}
\label{app:acquisition}
Table~\ref{tab:information_acquisition} includes all 74 initial decisions in the local-replacement process. Testing is suboptimal in 32 scenarios and optimal in 42. When testing is optimal, errors include skipping the test and testing the wrong channel. All models receive the same initial input for each scenario; subsequent histories depend on their choices and the resulting feedback.

The source-integration example is \texttt{U5.F10.T4.D}. Skipping has expected cost 10.008593; testing the unchanged channel has cost 10.262387, including the 0.30 fee. Both values assume optimal continuation. Figure~\ref{fig:reliance_results}c reports behavior frequencies over all scenarios in each subset; full decision breakdowns are listed in Table~\ref{tab:information_acquisition}.

\begin{table}[htbp]
\centering\footnotesize
\caption{First-stage decisions in the local-replacement process. Under skip-optimal scenarios (32 cases), testing is unnecessary. Under test-optimal scenarios (42 cases), failures include skipping the test or choosing suboptimal actions.}
\label{tab:information_acquisition}
\setlength{\tabcolsep}{4.5pt}
\renewcommand{\arraystretch}{1.1}
\begin{tabular}{@{}lrrrrrr@{}}
\toprule
& \multicolumn{3}{c}{Skip optimal (32 scenarios)} & \multicolumn{3}{c}{Test optimal (42 scenarios)}\\
\cmidrule(lr){2-4}\cmidrule(l){5-7}
Model & Skip & Test & Other errors & Optimal test & Skip & Other errors\\
\midrule
GPT-6 Luna & 19 & 13 & 0 & 36 & 4 & 2\\
GPT-5.6 Sol & 32 & 0 & 0 & 42 & 0 & 0\\
Gemini 3.8 Flash & 6 & 26 & 0 & 38 & 0 & 4\\
Claude Sonnet 5 & 3 & 29 & 0 & 38 & 2 & 2\\
Claude Haiku 4.5 & 11 & 21 & 0 & 30 & 8 & 4\\
DeepSeek V4.1 Flash & 22 & 5 & 5 & 36 & 2 & 4\\
Qwen3.8 27B & 15 & 14 & 3 & 33 & 5 & 4\\
\bottomrule
\end{tabular}
\end{table}

To assess whether errors occur only near the break-even point, we compute the expected benefit of each test before its fee. At the initial history $h_1$, for test $a$ with fee $c_a$, this is
\begin{equation}
B(a)=Q_1^*(h_1,\mathrm{skip})-\bigl[Q_1^*(h_1,a)-c_a\bigr].
\end{equation}
Both choices assume optimal continuation, so $B(a)$ measures the reduction in expected later costs attributable to that test. It can be zero even if the outcome changes beliefs about the partner. Values within $10^{-8}$ of zero are treated as numerical zero. In 17 scenarios, both tests have zero benefit; all are among the 32 skip-optimal scenarios. Across the seven models, 61 of 108 unnecessary test choices have zero benefit. This count also includes cases where the other test has a positive benefit below its fee. Table~\ref{tab:verification_value} distinguishes the 17-scenario subset from the individual tests chosen across all 32 skip-optimal scenarios.

\begin{table}[htbp]
\centering\footnotesize
\caption{Verification when information has no expected decision benefit. The 17-scenario subset contains cases where neither available test lowers expected later costs before its fee. The final column counts chosen tests with zero benefit among all unnecessary tests in the 32 skip-optimal scenarios; the two subsets differ.}
\label{tab:verification_value}
\setlength{\tabcolsep}{8pt}
\begin{tabular}{@{}lrr@{}}
\toprule
Model & \shortstack{Tests on zero-benefit\\scenarios (/17)} & \shortstack{Zero-benefit / all\\unnecessary tests}\\
\midrule
GPT-6 Luna & 5 & 6/13\\
GPT-5.6 Sol & 0 & --\\
Gemini 3.8 Flash & 12 & 15/26\\
Claude Sonnet 5 & 15 & 18/29\\
Claude Haiku 4.5 & 9 & 12/21\\
DeepSeek V4.1 Flash & 2 & 2/5\\
Qwen3.8 27B & 7 & 8/14\\
\bottomrule
\end{tabular}
\end{table}

\clearpage
\subsection{Learning through cooperation and explicit audits}
\label{app:information_planning}
The standard repeated-cooperation process contains 50 scenarios with stable partner modes, three production jobs, and quality feedback after each job. The 24 specialized scenarios are excluded from this comparison. For each initial history, let $M(h_1)$ be the set of actions minimizing immediate expected cost and $O(h_1)$ the set minimizing total expected cost under optimal continuation. We select cases where $M(h_1)\cap O(h_1)=\varnothing$, using a tolerance of $10^{-8}$ for numerical equality. This selects 15 scenarios across 13 activity categories independently of model responses. Costs scale by job size while the partner modes remain fixed, so feedback from earlier cooperation can inform later choices. All 15 scenarios remain in the denominators in Table~\ref{tab:information_planning}.

The route-planning example is \texttt{U5.F01.T1.A}. Reserving both routes has immediate expected cost 1.837 and cost 9.282070 under optimal continuation. Using the record adviser's route has immediate cost 1.860 and total cost 8.680600. The first choice changes which quality observation becomes available. In all 15 selected scenarios, every incorrect first action belongs to $M(h_1)$.

Seven of the selected blueprints have a corresponding delayed-feedback scenario in which an explicit audit is optimal. Ordinary execution in that workflow provides no first-job quality feedback; an audit adds 0.4 to the execution fee and supplies the check immediately. Production payoffs and initial mode probabilities are shared across workflows, while feedback timing and available actions differ. Table~\ref{tab:learning_audit_pairs} reports first-stage correctness on these seven blueprints. Figure~\ref{fig:partner_information}a includes all seven blueprints for each model.

\begin{table}[htbp]
\centering\small
\caption{First decisions on seven shared blueprints, each with an ordinary-cooperation and an explicit-audit scenario. Each column reports the number of optimal initial decisions out of seven. The blueprints are F01.T1, F05.T2, F07.T4, F09.T2, F10.T3, F11.T1, and F13.T3.}
\label{tab:learning_audit_pairs}
\setlength{\tabcolsep}{12pt}
\begin{tabular}{@{}lrr@{}}
\toprule
Model & \shortstack{Cooperation\\optimal (/7)} & \shortstack{Audit\\optimal (/7)}\\
\midrule
GPT-6 Luna & 4 & 5\\
GPT-5.6 Sol & 7 & 7\\
Gemini 3.8 Flash & 5 & 7\\
Claude Sonnet 5 & 0 & 7\\
Claude Haiku 4.5 & 0 & 1\\
DeepSeek V4.1 Flash & 3 & 5\\
Qwen3.8 27B & 1 & 7\\
\bottomrule
\end{tabular}
\end{table}

\Needspace{7\baselineskip}
\subsection{Assistance allocation}
\label{app:assistance}
The 74 improvement-or-assessment scenarios evaluate whether models correctly target support interventions, audit channels, or forego intervention. Beyond aggregate decision accuracy, characterizing assistance behavior requires examining specific failure modes across channels. Table~\ref{tab:assistance_allocation} separates errors that choose the wrong support target from other decision errors.

In the assembly example \texttt{U5.F06.T4.C}, both timing (P) and interface compatibility (Q) matter. Supporting either channel costs 0.6 and changes a low mode to a high mode with probability 0.75. Before support, the channel success probabilities are 0.690 for P and 0.550 for Q. Supporting P raises its probability to 0.8475, a gain of 0.1575; supporting Q raises its probability to 0.7750, a gain of 0.2250. The model must then choose how to perform two production jobs. Under optimal continuation, supporting P costs 12.38525 and supporting Q costs 14.31325. Sonnet nonetheless targets channel Q. This failure demonstrates that local, myopic reliability gains do not necessarily correspond to maximal downstream value in sequential collaboration.

\begin{table}[htbp]
\centering\footnotesize
\caption{Initial decisions in improvement-or-assessment scenarios (74 blueprints). A wrong support target means selecting support for one channel when supporting the other is optimal. Other decision errors include testing, skipping, or providing support when a different action type is optimal, as well as unparseable responses.}
\label{tab:assistance_allocation}
\setlength{\tabcolsep}{6pt}
\begin{tabular}{@{}lrrr@{}}
\toprule
Model & \shortstack{Optimal\\(/74)} & \shortstack{Wrong support\\target} & \shortstack{Other decision\\errors}\\
\midrule
GPT-6 Luna & 67 & 5 & 2\\
GPT-5.6 Sol & 74 & 0 & 0\\
Gemini 3.8 Flash & 67 & 5 & 2\\
Claude Sonnet 5 & 65 & 8 & 1\\
Claude Haiku 4.5 & 56 & 10 & 8\\
DeepSeek V4.1 Flash & 67 & 0 & 7\\
Qwen3.8 27B & 66 & 2 & 6\\
\bottomrule
\end{tabular}
\end{table}

\subsection{Reliability-support comparisons}
\label{app:reliability_support}
Table~\ref{tab:reliability_support} reports all-stage accuracy on the 24 specialized scenarios and decision changes under reliability support. The stage-level comparisons match preceding actions and feedback, and evaluate paired decisions where both responses are valid actions.

Qwen's number of scenarios with three optimal decisions rises from 12 to 20. Restricting to paired valid decisions leaves two improvements and no deteriorations among 39 comparisons.

\begin{table}[htbp]
\centering\small
\caption{Reliability support on 24 scenarios. Baseline and support columns count scenarios with optimal actions at all three stages. $+/-$ counts improved/worsened decisions among $n$ stage pairs. The final column evaluates paired decisions where both responses are valid actions.}
\label{tab:reliability_support}
\setlength{\tabcolsep}{3.5pt}
\renewcommand{\arraystretch}{1.1}
\begin{tabular}{@{}lrrrr@{}}
\toprule
Model & \shortstack{Baseline\\(/24)} & \shortstack{Support\\(/24)} & \shortstack{Same history\\$+/-$ ($n$)} & \shortstack{Valid pairs\\$+/-$ ($n$)}\\
\midrule
GPT-6 Luna & 18 & 19 & 4/3 (59) & 4/3 (59)\\
GPT-5.6 Sol & 24 & 23 & 0/1 (70) & 0/1 (70)\\
Gemini 3.8 Flash & 20 & 20 & 2/1 (67) & 2/1 (67)\\
Claude Sonnet 5 & 16 & 18 & 3/1 (64) & 2/0 (57)\\
Claude Haiku 4.5 & 16 & 17 & 6/5 (50) & 6/5 (46)\\
DeepSeek V4.1 Flash & 20 & 22 & 3/1 (66) & 2/0 (58)\\
Qwen3.8 27B & 12 & 20 & 8/0 (54) & 2/0 (39)\\
\bottomrule
\end{tabular}
\end{table}

\clearpage
\section{Limitations}
\label{app:limitations}

While the benchmark provides an analytically grounded framework for assessing calibrated reliance, several limitations contextualize our empirical findings and suggest directions for subsequent research.

\paragraph{Sequential horizon length and extended interactions.}
Each benchmark scenario evaluates a three-stage sequential decision process. This horizon length is chosen deliberately: three sequential decisions provide a well-controlled, analytically exact horizon that cleanly dissociates myopic safeguard adoption from forward-looking information gathering (e.g., incurring immediate verification or cooperation costs to inform subsequent reliance choices), while keeping the underlying partially observable Markov decision process analytically solvable and computationally tractable across hundreds of scenarios. Nonetheless, many real-world human--AI and multi-agent collaborations involve longer or indefinite interaction sequences. In extended interactions, additional behavioral dynamics arise, such as cumulative reputation formation, gradual partner drift over extended periods, and progressive renegotiation of collaborative responsibilities. Assessing how language models balance exploration, learning, and safeguard procurement across extended horizons represents a valuable direction for future work.

\paragraph{Single-trajectory evaluation and deterministic feedback draws.}
To guarantee strict, controlled comparability across models, each scenario evaluates a single trajectory under pre-sampled, fixed environmental feedback draws at each decision node. This design ensures that every model faces identical objective state transitions and diagnostic observations when taking identical actions, eliminating confounding variance from environmental sampling noise. Furthermore, our statistical confidence is grounded in the benchmark's breadth, spanning 324 scenarios (972 decision opportunities) across 74 distinct blueprints, with sampling uncertainty quantified through 10,000 bootstrap iterations and Wilson score intervals. Nonetheless, evaluating single trajectories under fixed draws does not capture generation-level stochasticity or model output variance across repeated trials and varied sampling temperatures. Future iterations could explore multi-trajectory rollouts and temperature sensitivity analyses to characterize within-model behavioral stability across stochastic runs.

\paragraph{Commercial API opacity and model evolution.}
Our benchmark evaluates frontier reasoning models accessed through commercial provider APIs. While commercial endpoints grant access to state-of-the-art capabilities, proprietary model services are subject to unannounced backend updates, provider-side prompt transformations, and behavioral drift over time. To ensure full reproducibility despite API opacity, the public benchmark release includes frozen records of the entire experimental workflow: the exact evaluation environments, input prompts, fixed pseudo-random feedback draws, raw model outputs, and automated scoring pipelines. These assets enable independent verification and re-scoring of all reported metrics without relying on active queries to external service endpoints.

\let\clearpage\ppclearpage
\end{document}